\documentclass[10pt,twocolumn,letterpaper]{article}

\usepackage{iccv}              

\definecolor{iccvblue}{rgb}{0.21,0.49,0.74}
\usepackage[pagebackref,breaklinks,colorlinks,allcolors=iccvblue]{hyperref}

\usepackage{graphicx}
\usepackage{amsmath}
\usepackage{amssymb}
\usepackage{amsfonts}
\usepackage{xcolor}
\usepackage{color}
\usepackage{colortbl}
\usepackage{soul}
\usepackage{makecell}
\newcommand{\modelname}{AD-GS}

\def\paperID{819} 
\def\confName{ICCV}
\def\confYear{2025}

\title{\modelname: Object-Aware B-Spline Gaussian Splatting for Self-Supervised Autonomous Driving}

\author{Jiawei Xu\renewcommand{\thefootnote}{\arabic{footnote}}\footnotemark[1]~~~
Kai Deng\renewcommand{\thefootnote}{\arabic{footnote}}\footnotemark[1]~~~
Zexin Fan\renewcommand{\thefootnote}{\arabic{footnote}}\footnotemark[1]~~~
Shenlong Wang\renewcommand{\thefootnote}{\arabic{footnote}}\footnotemark[2]~~~
Jin Xie\renewcommand{\thefootnote}{\arabic{footnote}}\footnotemark[3]\ ~\renewcommand{\thefootnote}{\fnsymbol{footnote}}\footnotemark[1]~~~
Jian Yang\renewcommand{\thefootnote}{\arabic{footnote}}\footnotemark[1]\ ~\renewcommand{\thefootnote}{\fnsymbol{footnote}}\footnotemark[1]\\
\renewcommand{\thefootnote}{\arabic{footnote}}\footnotemark[1] \ College of Computer Science, Nankai University~~
\renewcommand{\thefootnote}{\arabic{footnote}}\footnotemark[2] \ University of Illinois Urbana-Champaign \\
\renewcommand{\thefootnote}{\arabic{footnote}}\footnotemark[3] \ School of Intelligence Science and Technology, Nanjing University \\
{\tt\small \{jiaweixu, dengkai, zexin\_fan\}@mail.nankai.edu.cn \ 
\tt\small csjyang@nankai.edu.cn \ 
\tt\small csjxie@nju.edu.cn}
}

\begin{document}

\twocolumn[{%
\renewcommand\twocolumn[1][]{#1}%
\maketitle
\begin{center}
    \centering
    \captionsetup{type=figure}
    \includegraphics[width=\textwidth]{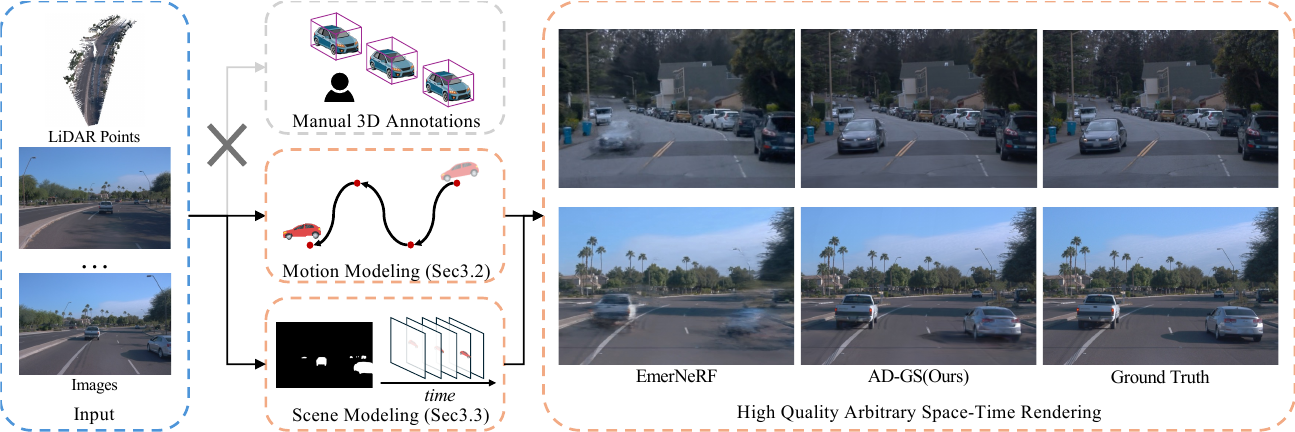}
    \captionof{figure}{\modelname. We achieve high-quality rendering by self-supervised manners for autonomous driving scenes \textbf{without} relying on the expensive manual 3D annotations.}
    \label{fig:teaser}
\end{center}%
}]

\renewcommand{\thefootnote}{\fnsymbol{footnote}}
\footnotetext[1]{Corresponding authors.}

\begin{abstract}
Modeling and rendering dynamic urban driving scenes is crucial for self-driving simulation. 
Current high-quality methods typically rely on costly manual object tracklet annotations, while self-supervised approaches fail to capture dynamic object motions accurately and decompose scenes properly, resulting in rendering artifacts. 
We introduce \modelname, a novel self-supervised framework for high-quality free-viewpoint rendering of driving scenes from a single log. 
At its core is a novel learnable motion model that integrates locality-aware B-spline curves with global-aware trigonometric functions, enabling flexible yet precise dynamic object modeling. 
Rather than requiring comprehensive semantic labeling, \modelname~automatically segments scenes into objects and background with the simplified pseudo 2D segmentation, representing objects using dynamic Gaussians and bidirectional temporal visibility masks. 
Further, our model incorporates visibility reasoning and physically rigid regularization to enhance robustness.
Extensive evaluations demonstrate that our annotation-free model significantly outperforms current state-of-the-art annotation-free methods and is competitive with annotation-dependent approaches. 
Project Page: \url{https://jiaweixu8.github.io/AD-GS-web/}
\end{abstract}

\section{Introduction}
\label{sec:intro}

Autonomous driving~(auto-driving) scene rendering models reconstruct dynamic driving environments using LiDAR points and images captured from multiple camera poses and timestamps.
These models enable rendering from novel viewpoints and moments.
Most existing approaches rely on manually annotated object bounding boxes and poses within the scenes to facilitate reconstruction~\cite{nsg, drivinggaussian, 4dgf, lihigs, mars, hierarchical, neurad, omnire, streetgs, unisim}.
Such 3D annotations simplify the process by eliminating the need for models to estimate object positions and motions, thereby making the reconstruction of large-scale auto-driving scenes much more easier.
While models based on manual 3D annotations have achieved significant progress in recent years, the high cost of annotation poses a potential barrier to their widespread adoption.

To address this challenge, self-supervised models for auto-driving scene rendering have been developed~\cite{emernerf, hugs, pvg, suds, rodus, splatflow, urban4d, desire-gs, pnf}.
These models aim to reconstruct both scenes and motions of the contained objects only using images and LiDAR points, thereby eliminating the reliance on manual 3D annotations.
However, their performance may still fall short of expectations.
Employing neural networks for motion modeling can be an effective approach but may result in high computational overhead and fail to capture the fine-grained local movements of objects~\cite{emernerf, splatflow, suds, rodus, urban4d, pnf}.
On the other hand, using predefined functions, such as trigonometric functions, to model motion can achieve high rendering speeds~\cite{pvg, desire-gs}.
Also, these functions can represent the general motions under the noisy self-supervision due to their global fitting properties which optimize all parameters for each training frame.
However, they might still have difficulties in handling the local details of motions.
For scene modeling, auto-driving scenes can be categorized into various components with distinct properties, such as houses, cars, and trees.
Previous research has approached scene segmentation for independent modeling using methods like unsupervised learning~\cite{splatflow}, feature-based self-supervision~\cite{emernerf, suds, desire-gs}, or fine-grained semantic self-supervision~\cite{urban4d, hugs, pnf, rodus}.
While these techniques are effective in scene decomposition, the use of the noisy pseudo ground truth can result in reconstruction artifacts.

In this paper, we introduce \modelname, a novel self-supervised model designed for high-quality auto-driving scene rendering based on Gaussian splatting~\cite{gaussian-splatting}.
To achieve both global and local motion fitting with high precision, we incorporate learnable B-spline curves and B-spline quaternion curves, combined with trigonometric functions, into the deformation representation of dynamic Gaussians.
Unlike the methods using only trigonometric functions~\cite{pvg, desire-gs}, B-spline curves fit local details by only optimizing the relevant control points instead of all for the corresponding training frames.
Additionally, \modelname~simplifies the scene modeling process by segmenting auto-driving scenes into two parts: objects and background, based on the simplified pseudo 2D segmentation.
For the object component, we represent objects using dynamic Gaussians and employ bidirectional temporal visibility masks to handle their sudden appearance or disappearance.
For the background component, we keep the Gaussians stationary.
This decomposition allows \modelname~to reconstruct each component more accurately, resulting in higher rendering quality under noisy pseudo ground truth conditions.
To ensure consistent deformation among Gaussians representing the same object, we introduce visibility and physically rigid regularization, which helps to prevent chaotic behavior.

We compare \modelname~with the state-of-the-art self-supervised auto-driving scene rendering models, and our experiments demonstrate the significant performance improvements achieved by \modelname.
Overall the contributions of this paper can be summarized as follows,
\begin{itemize}
    \item We propose a novel motion modeling method that represents the Gaussian deformation using learnable B-spline curves and B-spline quaternion curves, combined with trigonometric functions, for both local and global fitting.
    \item We propose a novel approach for modeling auto-driving scenes. Guided by the simplified pseudo 2D segmentation, our method segments a scene into two components: objects and background, ensuring robustness under noisy pseudo ground truth. Objects are reconstructed using dynamic Gaussians and bidirectional temporal visibility masks for enhanced accuracy.
    \item We design the visibility and physically rigid regularization, combined with self-supervision, to keep the model from chaotic behavior for better performance.
\end{itemize}

\section{Related Work}

\noindent\textbf{Manually Assisted Scene Rendering.}
NSG~\cite{nsg} applies Neural Radiance Field~(NeRF)~\cite{nerf} in a scene graph format, utilizing manual 3D annotations for the rendering of auto-driving scenes, while MARS~\cite{mars} introduces a modular design for this task.
NeuRAD~\cite{neurad} and UniSim~\cite{unisim} extend NeRF models to the modeling of LiDAR sensors and the development of safe self-driving vehicles.
While NeRF-based approaches for autonomous driving scenes have achieved significant progress, they often overlook pedestrians and other non-vehicle dynamic actors.
To address this limitation, OmniRe~\cite{omnire} utilizes SMPL~\cite{smpl} and deformation fields to model pedestrians and other dynamic entities.
Additionally, to better capture fast-moving objects, ML-NSG~\cite{hierarchical} introduces a multi-level formulation of NSG, achieving notable performance improvements.
With the advent of Gaussian splatting~\cite{gaussian-splatting}, many studies have applied Gaussian splatting to the rendering of the auto-driving scene~\cite{drivinggaussian, streetgs, 4dgf}, achieving high-quality results with reduced computational time.
4DGF~\cite{4dgf} integrates scene dynamics with NSG, yielding substantial advancements in urban reconstruction.
Despite the progress made by rendering models reliant on annotations, the high cost of manual annotations probably remains a barrier to their widespread adoption.

\noindent\textbf{Self-Supervised Scene Rendering.}
Self-supervised models reconstruct auto-driving scenes without the need for manual annotations, presenting a more challenging but highly promising alternative to manually assisted models.
PNF~\cite{pnf} leverages pseudo 3D annotations and jointly optimizes object poses for auto-driving scene rendering.
SUDS~\cite{suds} adopts a scene factorization approach, utilizing a three-branch hash table representation~\cite{instantngp} for 4D reconstruction.
To achieve higher rendering quality, EmerNeRF~\cite{emernerf} supervises encoded features with the detection models~\cite{dinov2} and performs static-dynamic decomposition.
For more robust static and dynamic element separation, RoDUS~\cite{rodus} uses 2D segmentation~\cite{mask2former} as a guide to improve the decomposition performance.
When it comes to Gaussian splatting, HUGS~\cite{hugs} integrates NSG and Gaussian splatting with pseudo labels for reconstruction, modeling motion using unicycle models.
PVG~\cite{pvg} demonstrates the effectiveness of capturing motion by modeling Gaussian deformation with trigonometric functions and periodic vibration.
Despite advances in Gaussian-based models through unsupervised learning~\cite{splatflow}, feature supervision~\cite{desire-gs}, and fine-grained pseudo labeling~\cite{urban4d}, their performance remains limited by the challenges posed by the noisy pseudo labels and the failure to capture local motion details.

\section{Methods}

\subsection{Overview}

We develop \modelname~using Gaussian splatting~\cite{gaussian-splatting}, a point-based rendering model primarily designed for object-level static scenes, offering high rendering quality.
The model represents a scene through multiple 3D Gaussians, where each Gaussian is defined by $G = \{\boldsymbol{\mu}, S, R, \sigma, \mathbf{c}\}$, representing position, scaling, rotation, opacity, and color, respectively.
The color is expressed using spherical harmonic coefficients~(SH).
These 3D Gaussians can be rendered and trained through differentiable rasterization~\cite{splatting, differentiable-splatting} with ground-truth images.
The Gaussians are first projected onto the camera planes using a view transform matrix $W$ and a projective Jacobian matrix $J$:
\begin{align}
    \Sigma' = JW\Sigma J^TW^T, \Sigma=RSS^TR^T.
\end{align}
Finally, a pixel $\mathbf{u}$ on the camera planes can be rendered by $N$ overlapped Gaussians using the $\alpha$-blending as the following formula:
\begin{align}
\mathbf{C}(\mathbf{u}) =& \sum_{i=1}^N \mathbf{c}_i \alpha_i \prod_{j=1}^{i-1}(1-\alpha_j), \notag\\ &\alpha_i = \sigma_i e^{-\frac{1}{2}(\mathbf{u}-\boldsymbol{\mu}^p_i)^T\Sigma'^{-1}_i(\mathbf{u}-\boldsymbol{\mu}^p_i)},
\end{align}
where $\boldsymbol{\mu}^P_i$ is the projected position of a Gaussian.
In optimization, the model~\cite{gaussian-splatting} involves pruning the low-opacity Gaussians and densifying the Gaussians based on the gradient and scaling through the splitting and cloning operations.

\begin{figure*}[ht]
    \centering
    \includegraphics[width=\linewidth]{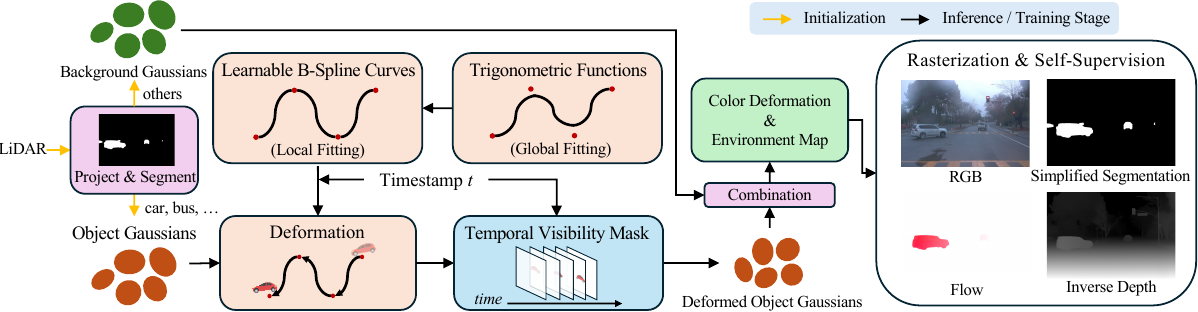}
    \caption{The pipeline of \modelname. We decompose the scene into two components: objects and background, and initialize the Gaussians with the simplified pseudo 2D segmentation. For the object Gaussians, we deform their parameters with learnable B-spline curves and trigonometric functions to achieve both local and global fitting. Additionally, we incorporate bidirectional temporal visibility masks to handle the sudden appearance and disappearance of the objects. In contrast, background Gaussians remain static. After integrating both components, we apply color deformation to all Gaussians and utilize a learnable environment map for distant regions. Finally, the rendering results are self-supervised using the pseudo labels and regularization.}
    \label{fig:pipeline}
\end{figure*}

To adapt Gaussian splatting for self-supervised auto-driving scene rendering, we divide all the $N$ Gaussians into two categories: objects $\boldsymbol{\Omega}_{obj}$ and background $\boldsymbol{\Omega}_{bkg}, |\boldsymbol{\Omega}_{obj}| + |\boldsymbol{\Omega}_{bkg}|=N$.
For the object Gaussians $G\in\boldsymbol{\Omega}_{obj}$, we model their motions by deforming their positions $\boldsymbol{\mu}$ and rotations $R$ over time using B-splines and trigonometric functions.
Additionally, we incorporate the bidirectional temporal visibility mask for these Gaussians to handle the sudden appearance and disappearance of the objects.
In contrast, the background Gaussians $G\in \boldsymbol{\Omega}_{bkg}$ remain stationary.
Notably, the colors of both object and background Gaussians may vary over time.
To account for this, similar to previous work~\cite{streetgs}, we deform the diffuse SH features $\mathbf{z}$ of all Gaussians $G\in \boldsymbol{\Omega}_{obj}\cup \boldsymbol{\Omega}_{bkg}$ using trigonometric functions:
\begin{align}\label{eqn:color}
\mathbf{z}' = \mathbf{z} + \sum_{l=1}^K \mathbf{f}_l\sin(l\pi\cdot t) + \mathbf{h}_l\cos(l\pi\cdot t).
\end{align}
$K$ is the maximum level of frequency, $\mathbf{f}_l$ and $\mathbf{h}_l$ are the learnable parameters, and $t$ is the timestamp.
Furthermore, we represent the distant parts of a scene, such as sky, using a learnable spherical environment map.
Our model is optimized in a self-supervised manner with regularization techniques to maintain stability and prevent chaotic behavior.
The pipeline of \modelname~is shown in Figure~\ref{fig:pipeline}.

\subsection{Learnable B-Spline Motion Curve}

\begin{figure}
    \centering
    \includegraphics[width=\linewidth]{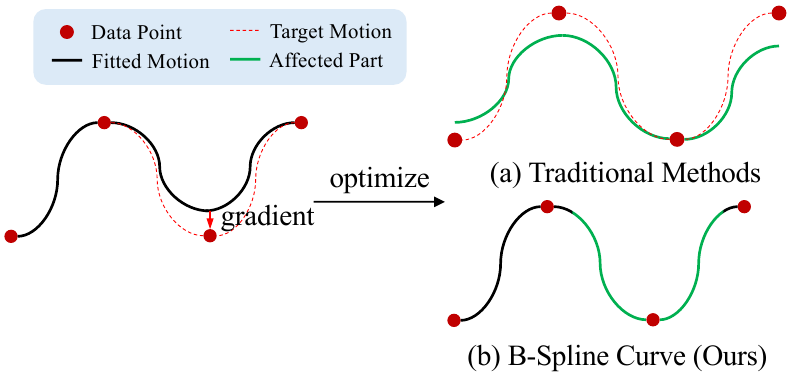}
    \caption{Comparison between traditional methods and the B-spline representation. (a) Traditional methods, such as MLPs and trigonometric functions, optimize all parameters globally to minimize the loss against the data points. This global optimization approach often limits their abilities to capture fine local details. (b) The B-spline curve achieves local detail fitting by only optimizing the nearby control points of a given data point, enabling more precise and flexible representation.}
    \label{fig:local-control}
\end{figure}

\noindent\textbf{Learnable B-Spline Curve.}
The B-spline curve is well known for its ability to provide local control.
Given $n + 1$ control points $\mathbf{p}_i, i\in \{0, 1, ..., n\}$, we can construct a $C^{k-1}$-continuous $k$-th order ($k \leq n + 1$) B-spline curve using the following formula,
\begin{align}\label{eqn:bspline}
    \mathbf{p}(t) = \sum_{i=0}^n \mathbf{p}_iB_{i, k}(t), t\in [t_{k - 1}, t_{n + 1}],
\end{align}
where $t_0, t_1, ..., t_{n+1}$ are the knots arranged in non-decreasing order.
In practice, we define $t$ as the timestamp, with $t_{k-1}$ and $t_{n+1}$ representing the minimum and maximum timestamps of the input sequence, respectively.
The basis function $B_{i, k}$ only has a non-zero value in $[t_i, t_{i+k}]$.
Furthermore, all control points $\mathbf{p}_i$ are treated as learnable parameters, and we utilize uniform B-spline curves for motion modeling.
However, the de Boor-Cox recursion formula for the basis function $B_{i, k}$ is computationally inefficient.
Therefore, we use the matrix formulation of uniform B-spline curves for better computational complexity~\cite{matbspline},
\begin{align}\label{eqn:matrix-basis}
    &[B_{i-k+1, k}(u), B_{i-k+2, k}(u) ..., B_{i, k}(u)] \notag \\ 
   =&[1, u, u^2, ..., u^{k-1}]M_k, \notag \\
   &u = \frac{t-t_i}{t_{i+1}-t_i}, u\in [0, 1), i=k-1, k, ..., n.
\end{align}
$M_k$ is a $k \times k$ matrix that can be precomputed before training, and the explicit expression can be found in Section~\ref{sec:supp-impl} of our supplementary.
Using Equation~\ref{eqn:matrix-basis}, we can obtain the matrix representation of the B-spline curve for each segment $[t_i, t_{i+1})$ in Equation~\ref{eqn:bspline}:
\begin{align}
    \mathbf{p}(t)=&[1, u, u^2, ..., u^{k-1}]M_k[\mathbf{p}_{i-k+1}, \mathbf{p}_{i-k+2}, ..., \mathbf{p}_{i}]^T \notag \\
    &u = \frac{t-t_i}{t_{i+1}-t_i}, u\in [0, 1), i=k-1, k, ..., n.
\end{align}
Unlike MLPs and trigonometric functions, which optimize all parameters for every output, B-spline curves exhibit local fitting. 
Each position on the B-spline curve is only influenced by the nearby control points, as illustrated in Figure~\ref{fig:local-control}. 
This property allows the B-spline curve to effectively capture motions with local details by only fine-tuning the relevant control points.
However, while the B-spline curve excels at modeling local details, it has limitations in capturing global motion, which is also crucial for accurate reconstruction under the noisy self-supervision.
To address this, we combine trigonometric functions with B-splines to model the movement of each object Gaussian $G\in \boldsymbol{\Omega}_{obj}$, achieving both global and local fitting:
\begin{align}\label{eqn:motion-full}
    \boldsymbol{\mu}' = \boldsymbol{\mu} + \mathbf{p}(t) + \sum_{l=1}^L \mathbf{a}_l\sin(t\cdot l\pi) + \mathbf{b}_l\cos(t\cdot l\pi),
\end{align}
where $L$ is the maximum frequency level, $\mathbf{a}_l$ and $\mathbf{b}_l$ are the learnable parameters.

\noindent\textbf{Learnable B-Spline Quaternion Curve.}
Because the interpolation of unit quaternions is uneven, the standard B-spline curve cannot adequately represent the unit quaternion deformation of Gaussian rotation parameters.
To address this limitation, we adopt the B-spline quaternion curve~\cite{quatbspline} to model Gaussian rotations.
Similar to Equation~\ref{eqn:bspline}, given $n + 1$ learnable unit quaternions $\mathbf{q}_i, i\in\{0, 1, ..., n\},$ as control points, the $k$-th order unit quaternion B-spline curve in each segment is defined as,
\begin{align}\label{eqn:quat-bspline}
    \mathbf{q}(t) =& \mathbf{q}_{i-k+1}\prod_{j=i-k+2}^i \exp(\mathbf{w}_j\widetilde{B}_{j, k}(t)), \notag \\
    &t\in [t_i, t_{i + 1}), i=k-1, k, ..., n,
\end{align}
where $\mathbf{w}_i=\log(\mathbf{q}_{i-1}^{-1}\mathbf{q}_i)$, and $\widetilde{B}_{i, k}(t) = \sum_{j=i}^n B_{j, k}(t)$, which can be efficiently computed using Equation~\ref{eqn:matrix-basis}.
Notably, the operators in Equation~\ref{eqn:quat-bspline} are specifically defined for unit quaternions.
Although the B-spline quaternion curve has limitations in global fitting, we find that directly modeling the rotation of each object Gaussian $G\in \boldsymbol{\Omega}_{obj}$ by setting $R=\mathbf{q}(t)$ is sufficient for our task.

\subsection{Object-Aware Splatting with Temporal Mask}
Having dynamic Gaussians deformed with B-spline curves for objects, \modelname~requires an accurate method to split and model a scene in objects and background.

\noindent\textbf{Scene Decomposition with Simplified 2D Segmentation.}
Previous works employ instance segmentation to guide scene decomposition~\cite{urban4d, rodus}, but this approach might perform excessively noisy.
Therefore, we simplify the segmentation classes into two categories: objects and background.
Classes likely to exhibit motion over time, such as cars, are categorized as objects, while all other classes are treated as background.
This simplification is sufficient and significantly more robust for rendering scenes in autonomous driving applications.
We initialize the Gaussians using LiDAR points and partition them into two subsets $\boldsymbol{\Omega}_{obj}$ and $\boldsymbol{\Omega}_{bkg}$, based on the 2D projected positions on the simplified binary segmentation results $\mathcal{M}_{obj}$ from SAM~\cite{sam, grounded-sam}.
Using this initial decomposition, we can render the object mask through the $\alpha$-blending by the following formula:
\begin{align}\label{eqn:obj-loss}
\hat{\mathcal{M}}_{obj} = \sum_{i=1}^N\mathbb{I}\{G_i\in \boldsymbol{\Omega}_{obj}\}\alpha_i\prod_{j=1}^{i-1}(1-\alpha_j),
\end{align}
where $\mathbb{I}$ denotes the indicator function.
The rendered object mask is supervised via the pseudo ground truth using the binary cross-entropy loss $\mathcal{L}_{obj} = BCE(\hat{\mathcal{M}}_{obj}, \mathcal{M}_{obj})$.
During training, $\mathcal{L}_{obj}$ makes the two Gaussian subsets stayed in their respective areas by adjusting their opacities.
More details can be found in Section~\ref{sec:supp-impl} of our supplementary.

\noindent\textbf{Bidirectional Temporal Visibility Mask.}
The moving objects might not be visible in all frames, necessitating a temporal mask to reduce the impact of invisible frames captured the already traversed places when the trajectories have not been fully fitted.
Therefore, we introduce the bidirectional temporal visibility mask applied to the opacity $\sigma$ of an object Gaussian $G\in\boldsymbol{\Omega}_{obj}$, defined by the formula:
\begin{align}\label{eqn:t-mask}
\sigma'(t) = \sigma \cdot e^{-\frac{(t-\mu_t)^2}{2s^2}}, 
s = \left\{ 
\begin{array}{cc}
     s_0, &t < \mu_t\\
     s_1, &t \geq \mu_t
\end{array}
\right.,
\end{align}
where $\mu_t$ is the fixed acquisition timestamp of the LiDAR points, $s_0, s_1$ are the learnable scales of the bidirectional visibility mask.
Unlike previous approaches~\cite{pvg}, which treat $\mu_t$ as a learnable parameter, we argue that the acquisition timestamp itself serves as an effective prompt to determine the visible moment of an object and distinguish the different objects that have passed through the same location.
To prevent the temporal mask from becoming excessively narrow, we design an expanding regularization loss as follows:
\begin{align}
\mathcal{L}_{s} = ||\frac{2\Delta_f}{s_0 + s_1}||_1,
\end{align}
where $\Delta_f$ denotes the average time interval between two consecutive frames along the input sequence.

In general, the learnable parameters for each object Gaussian are $G=\{\boldsymbol{\mu}, S, \sigma, \mathbf{c}, \mathbf{f}, \mathbf{h}, \mathbf{p}, \mathbf{a}, \mathbf{b}, \mathbf{q}, s_0, s_1\}, G\in\boldsymbol{\Omega}_{obj}$, while the learnable parameters for each background Gaussian are $G=\{\boldsymbol{\mu}, S, R, \sigma, \mathbf{c}, \mathbf{f}, \mathbf{h}\}, G\in\boldsymbol{\Omega}_{bkg}$.

\subsection{Self-Supervision with Regularization}

Unlike rendering models that rely on manual 3D annotations, self-supervised models require multiple pseudo ground truths to reconstruct large auto-driving scenes from sparse per-frame viewpoints.

\noindent\textbf{Flow Supervision.}
Flow supervision provides \modelname~with prompts regarding object motion, and we utilize CoTracker3~\cite{cotracker3} to generate the pseudo labels.
Following a similar approach to SUDS~\cite{suds}, we predict the 3D position of each pixel in the original image at the target moment defined by the pseudo ground truth:
\begin{align}
    \hat{X}_r = \sum_{i=1}^N\boldsymbol{\mu}_i'\alpha_i\prod_{j=1}^{i-1}(1-\alpha_j),
\end{align}
where $\boldsymbol{\mu}'$ is the position of each Gaussian at the target moment.
We then project the $\hat{X}_r$ onto the target image, and supervise it using the corresponding pseudo flow points $X_t$ through $\mathcal{L}_f = || \phi_t(\hat{X}_r) - X_t ||_1$, where $\phi_t$ is the projection function.
To enhance efficiency and reduce noise, we only supervise the flow of the object part with the simplified 2D pseudo segmentation results obtained from SAM~\cite{grounded-sam, sam}.

\noindent\textbf{Inverse Depth Supervision.}
In prior work, the depth map is rendered through the $\alpha$-blending of the distances between the camera and Gaussian centers, which is then inverted to supervise against the monocular pseudo ground truth.
However, when no Gaussian exists along a pixel ray, the inversion operation can become computationally unstable.
To address this problem, we directly render the expectation map of inverse depth~\cite{hierarchicalgs} by using the following formula:
\begin{align}
\frac{1}{\hat{d}} = \sum_{i=1}^N\frac{1}{d_i}\alpha_i\prod_{j=1}^{i-1}(1-\alpha_j),
\end{align}
where $d_i$ is the distance between camera and each Gaussian center.
Then we supervise the rendered inverse depth map with the monocular pseudo ground truth $\frac{1}{d}$ generated by DPTv2~\cite{dptv2}, using the scale-and-shift depth loss~\cite{monosdf}:
\begin{align}
    \mathcal{L}_d = ||w\frac{1}{\hat{d}} + q - \frac{1}{d}||_1.
\end{align}
$w, q$ are the optimal values that minimize $(w\frac{1}{\hat{d}} + q - \frac{1}{d})^2$.
This depth loss incorporates 3D information from sparse viewpoints, enabling \modelname~to reconstruct large auto-driving scenes more accurately.

\noindent\textbf{Visibility and Physically Rigid Regularization.}
To satisfy the assumption of local rigidity and prevent \modelname~from becoming disorganized, we introduce a regularization loss to ensure that nearby object Gaussians $G\in \boldsymbol{\Omega}_{obj}$ exhibit similar deformations in position and temporal visibility mask across the timeline:
\begin{align}\label{eqn:regularization}
    \mathcal{L}_r = \sum_{\boldsymbol{\beta}\in \{\mathbf{p}, \mathbf{a}, \mathbf{b}, \{s_0, s_1\}\}} \sum_{\beta_i\in \boldsymbol{\beta}}var(\beta_i).
\end{align}
$var$ means the parameter variance of the nearby 8 Gaussians selected by the KNN algorithm.
To address the high computational cost of the KNN operation, we cache the KNN results and update them every 10 iterations, striking a balance between efficiency and performance.

\noindent\textbf{Total Loss.}
In general, the loss functions for training \modelname~can be summarized as follows,
\begin{align}\label{eqn:total-loss}
    \mathcal{L} = &(1 - \lambda_c)\mathcal{L}_1 + \lambda_c\mathcal{L}_{D-SSIM} + \notag\\
    &\lambda_d\mathcal{L}_d + \lambda_f\mathcal{L}_f + \lambda_{obj}\mathcal{L}_{obj} + \lambda_{sky}\mathcal{L}_{sky} + \notag\\
    &\lambda_r\mathcal{L}_r + \lambda_s\mathcal{L}_s,
\end{align}
where $\lambda$ are the weight hyperparameters for each loss, $\mathcal{L}_1, \mathcal{L}_{D-SSIM}$ are the supervision using the ground truth images~\cite{gaussian-splatting}, and $\mathcal{L}_{sky} = BCE(1 - O, \mathcal{M}_{sky})$ is the binary cross-entropy loss between the accumulated opacity $O=\prod_{i=1}^{N}(1-\alpha_i)$ and the sky mask $\mathcal{M}_{sky}$ generated by SAM~\cite{sam, grounded-sam}.

\begin{table*}[!ht]
\centering
\small
\caption{Quantitative comparisons on the KITTI~\cite{kitti} dataset. We follow the experimental setup of SUDS~\cite{suds}, and the color of each cell shows the \sethlcolor{pink!150}\hl{best} and the \sethlcolor{orange!40}\hl{second best}. ``Annotations'' means whether the model is assisted by the manual 3D annotations.}
\setlength{\tabcolsep}{1.0mm}{
\begin{tabular}{lc|ccc|ccc|ccc}
\hline
& &\multicolumn{3}{c}{KITTI-$75\%$} & \multicolumn{3}{c}{KITTI-$50\%$} & \multicolumn{3}{c}{KITTI-$25\%$} \\
Model & Annotations & PSNR~$\uparrow$ & SSIM~$\uparrow$ & LPIPS~$\downarrow$ & PSNR~$\uparrow$ & SSIM~$\uparrow$ & LPIPS~$\downarrow$ & PSNR~$\uparrow$ & SSIM~$\uparrow$ & LPIPS~$\downarrow$ \\
\hline
\color{gray}StreetGS~\cite{streetgs} & \color{gray}$\checkmark$ & \color{gray}25.79 & \color{gray}0.844 & \color{gray}0.081 & \color{gray}25.52 & \color{gray}0.841 & \color{gray}0.084 & \color{gray}24.53 & \color{gray}0.824 & \color{gray}0.090 \\
\color{gray}ML-NSG~\cite{hierarchical} & \color{gray}$\checkmark$ & \color{gray}28.38 & \color{gray}0.907 & \color{gray}0.052 & \color{gray}27.51 & \color{gray}0.898 & \color{gray}0.055 & \color{gray}26.51 & \color{gray}0.887 & \color{gray}0.060 \\
\color{gray}4DGF~\cite{4dgf} & \color{gray}$\checkmark$ & \color{gray}31.34 & \color{gray}0.945 & \color{gray}0.026 & \color{gray}30.55 & \color{gray}0.931 & \color{gray}0.028 & \color{gray}29.08 & \color{gray}0.908 & \color{gray}0.036 \\
\hline
SUDS~\cite{suds} & & 22.77 & 0.797 & 0.171 & 23.12 & 0.821 & 0.135 & 20.76 & 0.747 & 0.198 \\
Grid4D~\cite{grid4d} & & 23.79 & 0.877 & 0.064 & 24.07 & 0.880 & 0.061 & 22.25 & \cellcolor{orange!40}0.846 & \cellcolor{orange!40}0.074 \\
PVG~\cite{pvg} & & \cellcolor{orange!40}27.13 & \cellcolor{orange!40}0.895 & \cellcolor{orange!40}0.049 & \cellcolor{orange!40}25.96 & \cellcolor{orange!40}0.885 & \cellcolor{orange!40}0.053 & \cellcolor{orange!40}22.59 & 0.841 & 0.078 \\
\modelname~(Ours) & & \cellcolor{pink!150}29.16 & \cellcolor{pink!150}0.920 & \cellcolor{pink!150}0.033 & \cellcolor{pink!150}28.51 & \cellcolor{pink!150}0.912 & \cellcolor{pink!150}0.035 & \cellcolor{pink!150}24.12 & \cellcolor{pink!150}0.868 & \cellcolor{pink!150}0.065 \\
\hline
\end{tabular}}
\label{tab:comp-kitti}
\end{table*}

\begin{figure*}[!ht]
    \centering
    \includegraphics[width=0.92\linewidth]{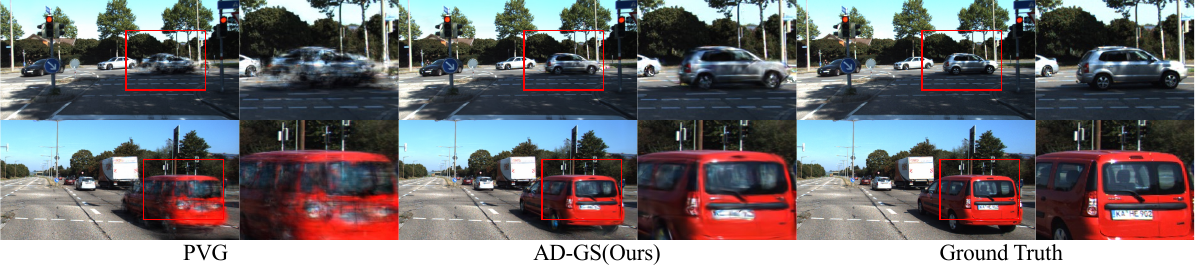}
    \caption{Qualitative comparisons on the KITTI~\cite{kitti} dataset.}
    \label{fig:comp-kitti}
\end{figure*}

\begin{table}[!ht]
\centering
\small
\caption{Quantitative comparisons on the Waymo~\cite{waymo} dataset. We mainly follow the experimental setup of StreetGS~\cite{streetgs} with a higher resolution $1280 \times 1920$, and the color of each cell shows the \sethlcolor{pink!150}\hl{best} and the \sethlcolor{orange!40}\hl{second best}. ``Annotations'' means whether the model is assisted by the manual 3D annotations. $^*$ denotes the metric only for moving objects.}
\setlength{\tabcolsep}{0.4mm}{
\begin{tabular}{lccccc}
\hline
Model & Annotations & PSNR~$\uparrow$ & SSIM~$\uparrow$ & LPIPS~$\downarrow$ & PSNR$^*$~$\uparrow$ \\
\hline
\color{gray}StreetGS~\cite{streetgs} & \color{gray}$\checkmark$ & \color{gray}33.97 & \color{gray}0.926 & \color{gray}0.227 & \color{gray}28.50 \\
\color{gray}4DGF~\cite{4dgf} & \color{gray}$\checkmark$ & \color{gray}34.64 & \color{gray}0.940 & \color{gray}0.244 & \color{gray}29.77 \\
\hline
PVG~\cite{pvg} & & 29.54 & 0.895 & 0.266 & 21.56 \\
EmerNeRF~\cite{emernerf} & & 31.32 & 0.881 & 0.301 & 21.80 \\
Grid4D~\cite{grid4d} & & \cellcolor{orange!40}32.19 & \cellcolor{orange!40}0.921 & \cellcolor{orange!40}0.253 & \cellcolor{orange!40}22.77 \\
\modelname~(Ours) & & \cellcolor{pink!150}33.91 & \cellcolor{pink!150}0.927 & \cellcolor{pink!150}0.228 & \cellcolor{pink!150}27.41 \\
\hline
\end{tabular}}
\label{tab:comp-waymo}
\end{table}

\begin{figure*}[!ht]
    \centering
    \includegraphics[width=0.92\linewidth]{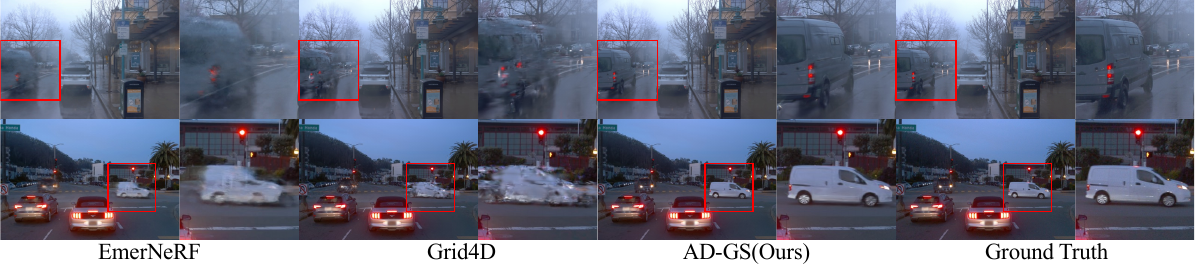}
    \caption{Qualitative comparisons on the Waymo~\cite{waymo} dataset.}
    \label{fig:comp-waymo}
\end{figure*}

\begin{table}[!ht]
\centering
\small
\caption{Quantitative comparisons on the nuScenes~\cite{nuscenes} dataset. We select six sequences with the resolution of $900\times1600$. The color of each cell shows the \sethlcolor{pink!150}\hl{best} and the \sethlcolor{orange!40}\hl{second best}.}
\setlength{\tabcolsep}{2mm}{
\begin{tabular}{lccc}
\hline
Model & PSNR~$\uparrow$ & SSIM~$\uparrow$ & LPIPS~$\downarrow$ \\
\hline
EmerNeRF~\cite{emernerf} & 27.17 & 0.853 & 0.268 \\
PVG~\cite{pvg} & 29.49 & 0.900 & 0.211 \\
Grid4D~\cite{grid4d} & \cellcolor{orange!40}30.29 & \cellcolor{orange!40}0.920 & \cellcolor{orange!40}0.172 \\
\modelname~(Ours) & \cellcolor{pink!150}31.06 & \cellcolor{pink!150}0.925 & \cellcolor{pink!150}0.164 \\
\hline
\end{tabular}}
\label{tab:comp-nuscenes}
\end{table}

\begin{figure*}[!ht]
    \centering
    \includegraphics[width=0.92\linewidth]{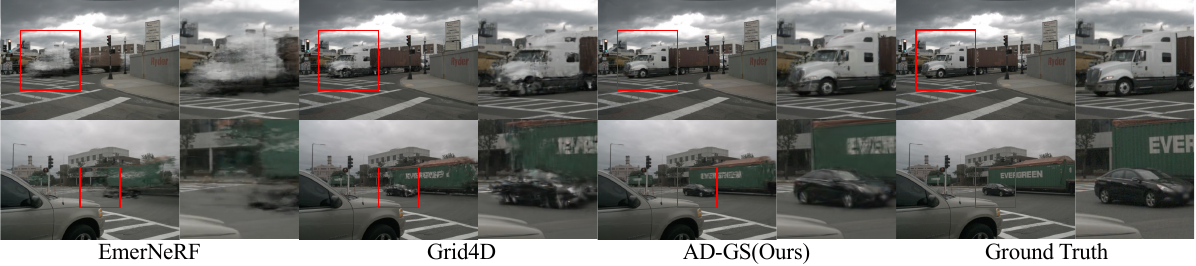}
    \caption{Qualitative comparisons on the nuScenes~\cite{nuscenes} dataset.}
    \label{fig:comp-nuscenes}
\end{figure*}

\section{Experiments}

\subsection{Setup}

\noindent\textbf{Datasets.}
We evaluate our model using three widely used datasets: KITTI~\cite{kitti}, Waymo~\cite{waymo} and nuScenes~\cite{nuscenes}.
For the KITTI dataset, we follow the settings of SUDS~\cite{suds} to conduct experiments on three sequences with the resolution of $375\times 1242$ captured by two cameras, where different proportions of images are used for training.
For the Waymo dataset, we mainly follow the settings of StreetGS~\cite{streetgs}, including eight sequences captured by a single camera, but we adopt a higher resolution of $1280\times 1920$.
For the nuScenes dataset, we select six sequences with three cameras and 60 frames, setting the resolution to $900\times 1600$.
We select one out of every four frames in the nuScenes sequences for testing, while the remaining frames are used for training.

\noindent\textbf{Baselines.}
We select several state-of-the-art self-supervised models to demonstrate the effectiveness of \modelname~for auto-driving scene rendering.
SUDS~\cite{suds} is a self-supervised model that employs a three-branch hash table representation for reconstruction.
EmerNeRF~\cite{emernerf} utilizes feature supervision from detection models to enhance self-supervised rendering.
PVG~\cite{pvg} is a Gaussian-based self-supervised model that represents object motions using trigonometric functions and periodic vibrations.
Due to the limited availability of more advanced self-supervised models specifically designed for auto-driving scene rendering, we also include Grid4D~\cite{grid4d}, a state-of-the-art model for object-level dynamic scene rendering, as a baseline.
To further evaluate our model, we additionally select several state-of-the-art models~\cite{streetgs, 4dgf, hierarchical} assisted by the manual 3D annotations.

\noindent\textbf{Implementation.}
We set $\lambda_c = 0.2, \lambda_d = 0.1, \lambda_f = 0.1, \lambda_{obj} = 0.1, \lambda_{sky} = 0.05, \lambda_r = 0.5$, and $\lambda_s = 0.01$ in Equation~\ref{eqn:total-loss} for all evaluations.
For the B-spline order $k$, we set $k = 6$ for all B-spline curves in Equation~\ref{eqn:bspline} and Equation~\ref{eqn:quat-bspline} across all datasets.
The number of control points for all B-splines $n + 1$ is set to one-third of the total number of frames in a sequence.
For the maximum level of trigonometric frequency, we set $K = L = 6$ in Equation~\ref{eqn:color} and Equation~\ref{eqn:motion-full} for all datasets.
We follow a training schedule and densification strategy similar to that of the original Gaussian splatting~\cite{gaussian-splatting}.
All experiments are conducted on a single RTX 3090 GPU.
Additional details can be found in Section~\ref{sec:supp-impl} and Section~\ref{sec:supp-setup} of our supplementary.

\subsection{Comparison}

\noindent\textbf{Comparison with Self-Supervised Models.}
We compare \modelname~with the state-of-the-art self-supervised models on the KITTI~\cite{kitti}, Waymo~\cite{waymo} and nuScenes~\cite{nuscenes} datasets.
The results are presented in Table~\ref{tab:comp-kitti} and Figure~\ref{fig:comp-kitti} for KITTI, Table~\ref{tab:comp-waymo} and Figure~\ref{fig:comp-waymo} for Waymo, and Table~\ref{tab:comp-nuscenes} and Figure~\ref{fig:comp-nuscenes} for nuScenes.
Notably, all the LPIPS~\cite{lpips} evaluations on the KITTI dataset are based on AlexNet~\cite{alexnet}, while the evaluations on the other datasets use VGG~\cite{vgg}.
Experimental results across all three datasets demonstrate that our model significantly outperforms the state-of-the-art self-supervised models.
Compared to PVG~\cite{pvg}, which relies solely on trigonometric functions for motion representation, our method benefits from the local fitting properties of learnable B-spline curves, as illustrated in Figure~\ref{fig:comp-kitti}.
Additionally, as shown in Figures~\ref{fig:comp-waymo} and~\ref{fig:comp-nuscenes}, the object-aware splatting technique in \modelname, including the simplified 2D segmentation and the bidirectional temporal visibility masks, achieves higher rendering quality than EmerNeRF~\cite{emernerf}, which uses feature supervision to help scene decomposition.
More results can be found in Section~\ref{sec:supp-exp} of our supplementary.



\noindent\textbf{Comparison with Manually Assisted Models.}
As shown in Table~\ref{tab:comp-kitti} and Table~\ref{tab:comp-waymo}, \modelname~achieves comparable performance despite not relying on manual 3D annotations for motion and scene reconstruction.
These results highlight the good performance of \modelname~in the self-supervised scene rendering.
However, when the training views become too sparse, such as the setting ``KITTI-25\%" in Table~\ref{tab:comp-kitti} which only uses 25\% images for training, self-supervised models might have difficulties in fitting the correct object motions by themselves.
However, our model still achieves improvements compared to self-supervised models under the extremely sparse view setting.

\subsection{Ablation Study}

\begin{table}[!ht]
\centering
\small
\caption{Loss ablation on the KITTI~\cite{kitti} dataset. The color of each cell shows the \sethlcolor{pink!150}\hl{best} and the \sethlcolor{orange!40}\hl{second best}.}
\setlength{\tabcolsep}{1.2mm}{
\begin{tabular}{ccc|ccc}
\hline
obj\&sky & flow\&depth & reg & PSNR~$\uparrow$ & SSIM~$\uparrow$ & LPIPS~$\downarrow$ \\
\hline
& & & 26.52 & 0.896 & 0.053 \\
$\checkmark$ & & & 26.98 & 0.902 & 0.048 \\
$\checkmark$ & $\checkmark$ & & \cellcolor{orange!40}28.03 & \cellcolor{orange!40}0.910 & \cellcolor{orange!40}0.042 \\
$\checkmark$ & $\checkmark$ & $\checkmark$ & \cellcolor{pink!150}29.16 & \cellcolor{pink!150}0.920 & \cellcolor{pink!150}0.033 \\
\hline
\end{tabular}}
\label{tab:abl-loss}
\end{table}

\begin{figure*}[!ht]
    \centering
    \includegraphics[width=0.95\linewidth]{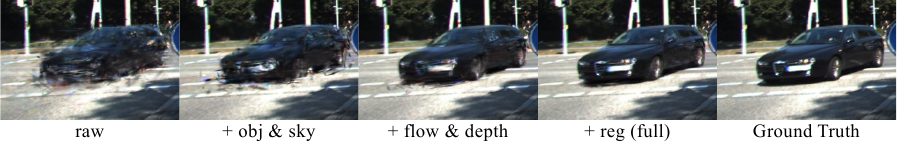}
    \caption{Visualization of loss ablation on the KITTI~\cite{kitti} dataset by gradually adding the losses.}
    \label{fig:abl-loss}
\end{figure*}

\noindent\textbf{Ablation of Losses.}
We evaluate the impact of each loss term under the ``KITTI-75\%" setting by progressively adding $\mathcal{L}_{obj}\&\mathcal{L}_{sky}$, $\mathcal{L}_f\&L_d$ and $\mathcal{L}_r$ in Equation~\ref{eqn:total-loss}, corresponding to the ``obj\&sky", ``flow\&depth" and ``reg", respectively.
As shown in Table~\ref{tab:abl-loss} and Figure~\ref{fig:abl-loss}, the 2D segmentation guidance provided by $\mathcal{L}_{obj}\&\mathcal{L}_{sky}$ helps the object and background Gaussians adjust their positions, leading to clearer rendering.
Additionally, $\mathcal{L}_d\&\mathcal{L}_f$ contribute extra motion and 3D information, enabling \modelname~to reconstruct the scene more accurately.
The regularization term $\mathcal{L}_r$ encourages \modelname~to maintain structural coherence, preventing disorganization and reducing artifacts.

\begin{table}[!ht]
\centering
\small
\caption{Object modeling module ablation on the Waymo~\cite{waymo} dataset. The color of each cell shows the \sethlcolor{pink!150}\hl{best} and the \sethlcolor{orange!40}\hl{second best}. $^*$ denotes the metric only for moving objects.}
\setlength{\tabcolsep}{0.7mm}{
\begin{tabular}{ccc|cccc}
\hline
sin\&cos & B-spline & t-mask & PSNR$^*$~$\uparrow$ & PSNR~$\uparrow$ & SSIM~$\uparrow$ & LPIPS~$\downarrow$ \\
\hline
 & $\checkmark$ & & 24.28 & 32.61 & 0.922 & 0.234 \\
$\checkmark$ & & & 25.70 & 33.38 & 0.925 & 0.232  \\
$\checkmark$ & $\checkmark$ & & \cellcolor{orange!40}26.65 & \cellcolor{orange!40}33.65 & \cellcolor{orange!40}0.926 & \cellcolor{orange!40}0.231 \\
$\checkmark$ & $\checkmark$ & $\checkmark$ & \cellcolor{pink!150}27.41 & \cellcolor{pink!150}33.91 & \cellcolor{pink!150}0.927 & \cellcolor{pink!150}0.228 \\
\hline
\end{tabular}}
\label{tab:abl-module}
\end{table}

\begin{figure}[!ht]
    \centering
    \includegraphics[width=0.95\linewidth]{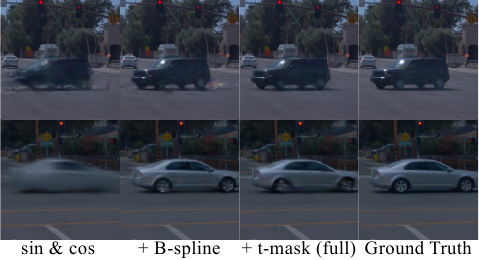}
    \caption{Visualization of object modeling module ablation on the Waymo~\cite{waymo} dataset by gradually adding the modules.}
    \label{fig:abl-module}
\end{figure}

\noindent\textbf{Ablation of Modules.}
To demonstrate the effectiveness of each module in \modelname~for object modeling, we perform an ablation study by progressively incorporating the learnable B-spline curves and the bidirectional temporal visibility masks, corresponding to ``B-spline" and ``t-mask", respectively.
For the ``sin\&cos" setting, the B-spline quaternion curves are replaced by the trigonometric functions.
The results are presented in Table~\ref{tab:abl-module} and Figure~\ref{fig:abl-module}.
Only using B-splines can result in over-fitting due to the noisy self-supervision, and trigonometric functions reduce the noise through their global optimization properties.
The last row of Figure~\ref{fig:abl-module} illustrates the scenarios in which a car suddenly appears and disappears.
Although the trigonometric functions successfully capture the general motion under the noisy self-supervision, the representation is inaccurate due to the incorrect optimization during the nearby training frames where the car is invisible.
The B-spline curves provide local motion fitting for more accurate representation to reduce the negative effect from the invisible frames, and the bidirectional temporal visibility masks further mitigate this problem.
More details and ablation studies can be found in Section~\ref{sec:supp-exp} of our supplementary.

\section{Conclusion}
In this paper, we introduced \modelname, a self-supervised Gaussian-based model for high-quality auto-driving scene rendering without relying on the manual 3D annotations.
\modelname~utilizes the learnable B-spline curves and B-spline quaternion curves, combined with trigonometric functions, to provide both local and global motion fitting. 
To achieve more accurate and robust scene modeling, we decomposed each scene into objects and background based on the simplified pseudo 2D segmentation results. 
The objects are represented by dynamic Gaussians with bidirectional temporal visibility masks to handle sudden appearances and disappearances, while the background remains static.
In addition to the self-supervision with pseudo labels, we proposed the visibility and physically rigid regularization techniques to prevent chaotic behavior in \modelname. 
Our experiments demonstrate the state-of-the-art performance achieved by \modelname~under self-supervised conditions.
However, achieving better rendering quality than annotation-dependent models still needs further research.

\clearpage


{
    \small
    \bibliographystyle{ieeenat_fullname}
    \bibliography{main}
}

\clearpage
\setcounter{page}{1}
\maketitlesupplementary
\appendix

\section{Implementation Details}\label{sec:supp-impl}

\noindent\textbf{Details in Equation~\ref{eqn:matrix-basis}.}
According to the prior work~\cite{matbspline}, the explicit expression of $M_k$ is
\begin{align}
M_k = \frac{1}{k-1}&\left(\left[\begin{array}{c}
     M_{k-1}\\
     0
\end{array}\right]\left[\begin{array}{ccccc}
     1 & k-2 &        &        & 0 \\
       & 2   & k-3    &        &   \\
       &     & \ddots & \ddots &   \\
     0 &     &        & k-1    & 0
\end{array}\right]\right.\notag\\ &+ \left.\left[\begin{array}{c}
     0 \\
     M_{k-1} 
\end{array}\right]\left[\begin{array}{ccccc}
    -1 & 1   &        &        & 0 \\
       & -1  & 1      &        &   \\
       &     & \ddots & \ddots &   \\
    0  &     &        & -1     & 1
\end{array}\right]\right),\notag\\
M_1=[1].
\end{align}
The above expression shows that $M_k$ can be precomputed before training to improve efficiency.

\noindent\textbf{Simplified Pseudo 2D Segmentation.}
We use Grounded-SAM-2~\cite{grounded-sam, sam, grounding-dino} with the Grounding DINO~\cite{grounding-dino} base model to generate the segmentation results as the object mask.
For the object segmentation in the KITTI~\cite{kitti} and Waymo~\cite{waymo} datasets, we use the prompt ``car.bus.truck.van.human".
Additionally, for the nuScenes~\cite{nuscenes} dataset, we include ``bike" as an extra prompt.
To generate pseudo labels for the sky mask, we use the prompt ``sky".

\noindent\textbf{Attribute Inheritance in Densification.}
In \modelname, the object Gaussians can only generate object Gaussians through the splitting or cloning operations, which are the same as the background Gaussians.
The newly created Gaussians will inherit all parameters from their parents, including the fixed parameter $\mu_t$ in Equation~\ref{eqn:t-mask}.
With this design, the loss $\mathcal{L}_{obj}$ in Equation~\ref{eqn:obj-loss} optimizes the opacity of each Gaussian, allowing object or background Gaussians with low opacities in incorrect locations to be pruned, thereby refining their numbers and positions.

\noindent\textbf{Others.}
Following StreetGS~\cite{streetgs}, we incorporate Structure-from-Motion~(SfM)~\cite{colmap} points as the initial background Gaussians to account for regions beyond the LiDAR scan range.
To mitigate the impact of imprecise camera poses in the KITTI and nuScenes datasets, we use a unified deformation for the positions of all Gaussians $G\in\Omega_{obj}\bigcup\Omega_{bkg}$.
Additionally, for the learnable spherical environment map, we set the resolution to $8192\times8192$.

\section{Experimental Setup Details}\label{sec:supp-setup}

\noindent\textbf{Dataset.}
For the KITTI~\cite{kitti} dataset, we select 0001, 0002 and 0006 sequences for evaluation with the left and right cameras.
For the Waymo~\cite{waymo} dataset, we select seg104481, seg123746, seg176124, seg190611, seg209468, seg424653, seg537228 and seg839851 with the FRONT camera, and use one out of every four frames for testing.
For the nuScenes~\cite{nuscenes} dataset, we select 0230, 0242, 0255, 0295, 0518 and 0749 scenes from 10 to 69(inclusive) frames with the FRONT, FRONT\_LEFT, and FRONT\_RIGHT cameras.
 
\noindent\textbf{Baselines.}
We mainly use the official implementation of EmerNeRF~\cite{emernerf} for the experiments on the Waymo and nuScenes dataset.
For the PVG~\cite{pvg}, we apply the hyperparameters designed for the Waymo dataset to evaluate its performance on the nuScenes dataset.
Notably, we use the same sky masks generated by SAM~\cite{grounded-sam, sam, grounding-dino} for EmerNeRF, PVG and \modelname~(Ours).
To adapt Grid4D~\cite{grid4d} for auto-driving scenarios, we extend it by increasing the temporal grid resolution to $1024\times1024\times32$.
We adapt 4DGF~\cite{4dgf} for our experimental setting on the Waymo dataset by adding the SfM points and disable the camera optimization.

\noindent\textbf{Others.}
We use the dynamic mask from StreetGS~\cite{streetgs} to compute the PSNR$^*$ only for moving objects on the Waymo dataset in Table~\ref{tab:comp-waymo} and Table~\ref{tab:abl-module}. 

\section{Additional Results}\label{sec:supp-exp}

\begin{table}[t]
\centering
\small
\caption{Ablation of the B-spline control points on the KITTI~\cite{kitti} dataset by changing the ratio between the number of control points and total frames. The color of each cell shows the \sethlcolor{pink!150}\hl{best} and the \sethlcolor{orange!40}\hl{second best}.}
\setlength{\tabcolsep}{2mm}{
\begin{tabular}{lccc}
\hline
Frames per Ctrl Pts & PSNR~$\uparrow$ & SSIM~$\uparrow$ & LPIPS~$\downarrow$ \\
\hline
1 & 27.26 & 0.902 & 0.046 \\
2 & 28.69 & \cellcolor{orange!40}0.910 & \cellcolor{orange!40}0.038 \\
3~(Ours) & \cellcolor{pink!150}29.16 & \cellcolor{pink!150}0.920 & \cellcolor{pink!150}0.033 \\
4 & \cellcolor{orange!40}29.11 & \cellcolor{pink!150}0.920 & \cellcolor{pink!150}0.033 \\
\hline
\end{tabular}}
\label{tab:supp-abl-bspline-ctrl}
\end{table}

\begin{table}[t]
\centering
\small
\caption{Ablation of the B-spline order on the KITTI~\cite{kitti} dataset. The color of each cell shows the \sethlcolor{pink!150}\hl{best}.}
\setlength{\tabcolsep}{2mm}{
\begin{tabular}{lccc}
\hline
Order & PSNR~$\uparrow$ & SSIM~$\uparrow$ & LPIPS~$\downarrow$ \\
\hline
$k=2$ & 29.10 & 0.919 & \cellcolor{pink!150}0.033 \\
$k=6$~(Ours) & \cellcolor{pink!150}29.16 & \cellcolor{pink!150}0.920 & \cellcolor{pink!150}0.033 \\
$k=10$ & 29.12 & \cellcolor{pink!150}0.920 & \cellcolor{pink!150}0.033 \\
$k=6$ + quat sin\&cos & 28.99 & 0.919 & 0.034 \\
\hline
\end{tabular}}
\label{tab:supp-abl-bspline-order}
\end{table}

\begin{figure}[t]
    \centering
    \includegraphics[width=\linewidth]{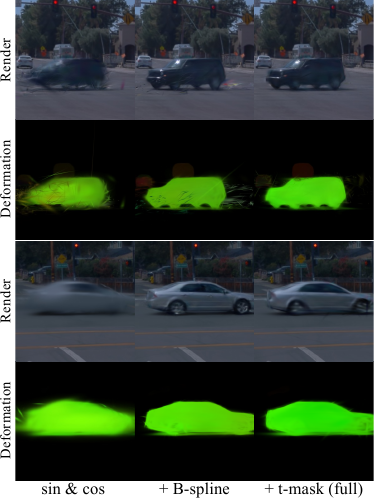}
    \caption{Deformation map of the object modeling module ablation study. In the deformation map, similar colors indicate similar deformations, and B-spline curves enhance the clarity and accuracy of the visualization. The bottom scenario depicts a car that is visible only briefly. In this case, optimization might be influenced by incorrect gradients from the training frames where the car is invisible. The results further demonstrate the effectiveness of B-spline curves in local fitting.}
    \label{fig:supp-deform}
\end{figure}

\begin{table}[t]
\centering
\small
\caption{Rendering speed comparison on the KITTI~\cite{kitti} dataset with self-supervised models. The color of each cell shows the \sethlcolor{pink!150}\hl{best} and the \sethlcolor{orange!40}\hl{second best}.}
\setlength{\tabcolsep}{2mm}{
\begin{tabular}{lccc}
\hline
Model & Grid4D~\cite{grid4d} & PVG~\cite{pvg} & \modelname~(Ours) \\
\hline
PSNR~$\uparrow$ & 23.79 & \cellcolor{orange!40}27.13 & \cellcolor{pink!150}29.16 \\
FPS~$\uparrow$ & 40 & \cellcolor{pink!150}58 & \cellcolor{orange!40}47 \\
\hline
\end{tabular}}
\label{tab:supp-speed}
\end{table}

\begin{figure}[t]
    \centering
    \includegraphics[width=\linewidth]{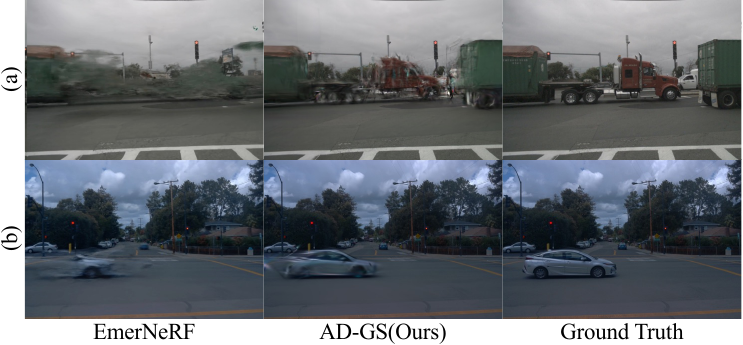}
    \caption{Failure cases when facing complex objects~(a) and objects only visible in a quite short time~(b).}
    \label{fig:supp-limitations}
\end{figure}

\begin{table}[h]
\centering
\small
\caption{Quantitative comparison by only removing flow supervision~$L_f$ in our model. The color of each cell shows the \sethlcolor{pink!150}\hl{best} and the \sethlcolor{orange!40}\hl{second best}. $^*$ denotes the metric only for moving objects.}
\setlength{\tabcolsep}{0.6mm}{
\begin{tabular}{l|cccc|ccc}
    \hline
    & \multicolumn{4}{c}{Waymo} & \multicolumn{3}{c}{KITTI-75\%} \\
    Model & PSNR & SSIM & LPIPS & PSNR$^*$ & PSNR & SSIM & LPIPS \\
    \hline
    PVG & 29.54 & 0.895 & 0.266 & 21.56 & 27.13 & 0.895 & 0.049  \\
    EmerNeRF & 31.32 & 0.881 & 0.301 & 21.80 & - & - & - \\
    Ours~w/o$L_f$ & \cellcolor{orange!40}33.20 & \cellcolor{orange!40}0.925 & \cellcolor{orange!40}0.229 & \cellcolor{orange!40}25.32 & \cellcolor{orange!40}28.84 & \cellcolor{orange!40}0.917 & \cellcolor{orange!40}0.036 \\
    Ours & \cellcolor{pink!150}33.91 & \cellcolor{pink!150}0.927 & \cellcolor{pink!150}0.228 & \cellcolor{pink!150}27.41 & \cellcolor{pink!150}29.16 & \cellcolor{pink!150}0.920 & \cellcolor{pink!150}0.033 \\
    \hline
\end{tabular}}
\label{tab:supp-abl-flow}
\end{table}

\begin{figure}[t]
    \centering
    \includegraphics[width=\linewidth]{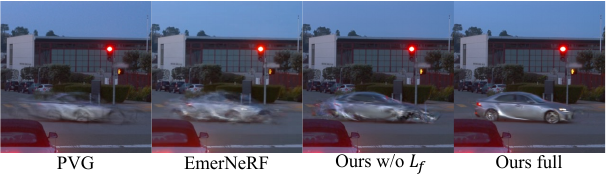}
    \caption{Qualitative comparison by removing flow supervision in our model.}
    \label{fig:supp-abl-flow}
\end{figure}

\noindent\textbf{Ablation of B-Splines.}
We conduct additional ablation studies on the parameters of B-spline curves using the ``KITTI-75\%" setting.
The results are presented in Table~\ref{tab:supp-abl-bspline-order} for the order $k$ and Table~\ref{tab:supp-abl-bspline-ctrl} for the number of control points.
When the order is too low or the control points are overly dense, the smoothness of the B-spline curves is constrained.
In this case, each control point is optimized using fewer frames, leading to performance degradation under noisy self-supervision.
Conversely, when the order is too high or the control points is insufficient, the local fitting capability decreases, resulting in artifacts.
Based on these observations, we select an order of $k = 6$ and set the ratio between the number of control points and the total frames to $1/3$ in our experiments to balance the smoothness and the local fitting property.
Additionally, we perform an ablation study on the deformation of the rotation parameter, with results shown in the last row of Table~\ref{tab:supp-abl-bspline-order}.
The setting ``sin\&cos" refers to modeling Gaussian rotation deformations using trigonometric functions, and the results show that the trigonometric functions are not necessary for the rotation parameters.

\noindent\textbf{Ablation of Optical Flow Supervision.}
A certain optical flow supervision is essential for this task, as it significantly aids in reconstructing the fast-moving objects commonly found in auto-driving scenarios. 
Pixels corresponding to such objects often exhibit significant displacements over time, making accurate matching challenging in the absence of flow supervision for trajectory reconstruction.
We conducted an additional experiment in which only the flow supervision term $L_f$ of our model was removed, and the results are shown in Table~\ref{tab:supp-abl-flow}.
Although our model still outperforms previous methods without optical flow, Figure~\ref{fig:supp-abl-flow} exhibits the obvious degradation caused by the absence of flow supervision, particularly in cases with fast-moving cars crossing the scene.

\noindent\textbf{Analysis of Motion Fitting.}
To further demonstrate the effectiveness of trigonometric function and B-spline curve in global and local fitting, we visualize the deformation map mainly following the approach in Grid4D~\cite{grid4d}.
The results are shown in Figure~\ref{fig:supp-deform}, where similar colors indicate similar deformations.
Although trigonometric functions can approximate the general deformation of an object under the noisy self-supervision, their representation tends to be inaccurate due to the omission of per-frame local details.
In contrast, B-spline curves offer advantages in capturing local details, allowing for more precise fine-tuning of the representation.
The bottom scenario in Figure~\ref{fig:supp-deform} illustrates a special case where a car suddenly appears and then disappears, remaining visible for only about 17 frames~(total about 160 frames).
When the model has not been fully optimized, the car still appears in the invisible frames.
However, the invisible frames cannot provide the correct information for the model to fit the trajectory at their timestamps.
In such cases, trigonometric functions might be influenced by numerous invisible training frames, leading to incorrect gradients during optimization and resulting in severe blurring in motion representation and rendering.
However, B-spline curves mitigate this issue by optimizing only the relevant control points, thereby reducing the impact of incorrect gradients from invisible training frames and significantly improving the accuracy of the representation.
Therefore, by combining trigonometric functions and B-spline curves for motion fitting, we achieve more accurate motion representations.

\noindent\textbf{Rendering Speed.}
We evaluate the rendering speed of \modelname~on the KITTI dataset with the ``KITTI-75" setting.
As shown in Table~\ref{tab:supp-speed}, \modelname~maintains fast rendering performance while improving quality, benefiting from the low computational overhead of trigonometric functions and B-spline curves.

\begin{figure*}[t]
    \centering
    \includegraphics[width=\linewidth]{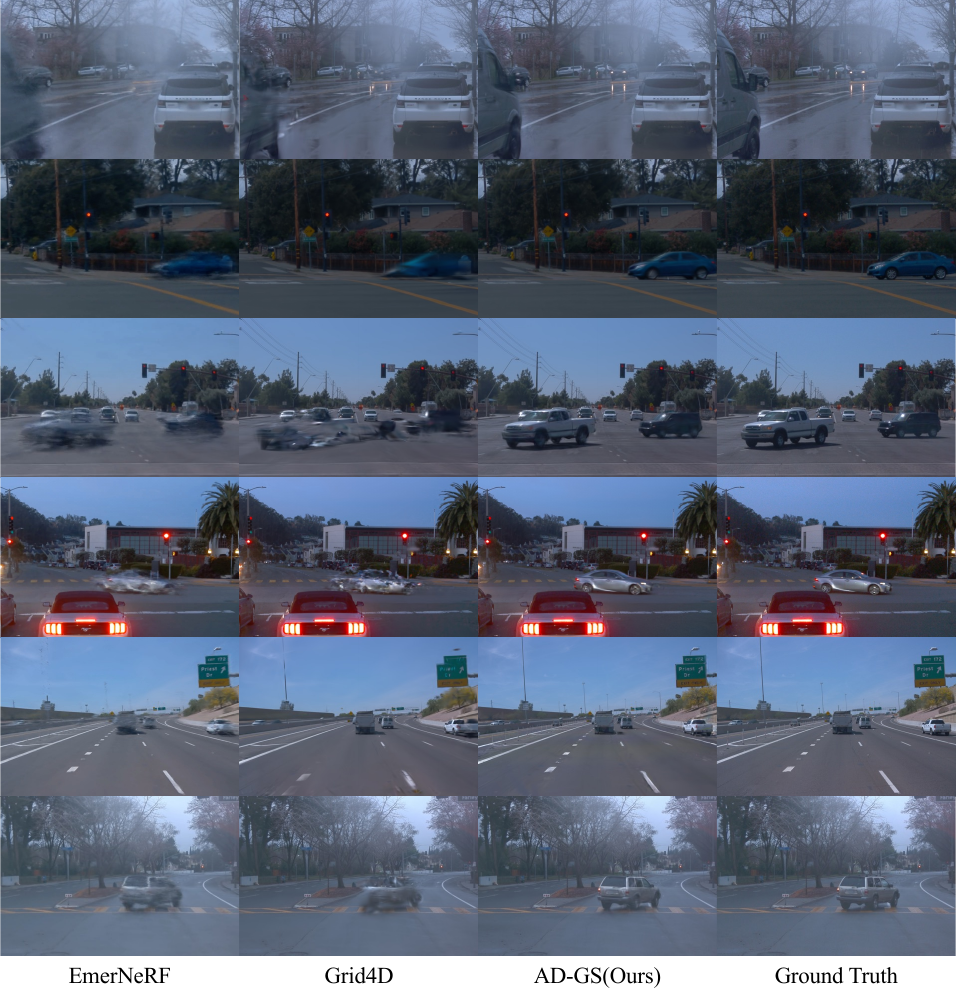}
    \caption{Additional qualitative comparisons on the Waymo~\cite{waymo} dataset.}
    \label{fig:supp-comp-waymo}
\end{figure*}

\begin{figure*}[t]
    \centering
    \includegraphics[width=\linewidth]{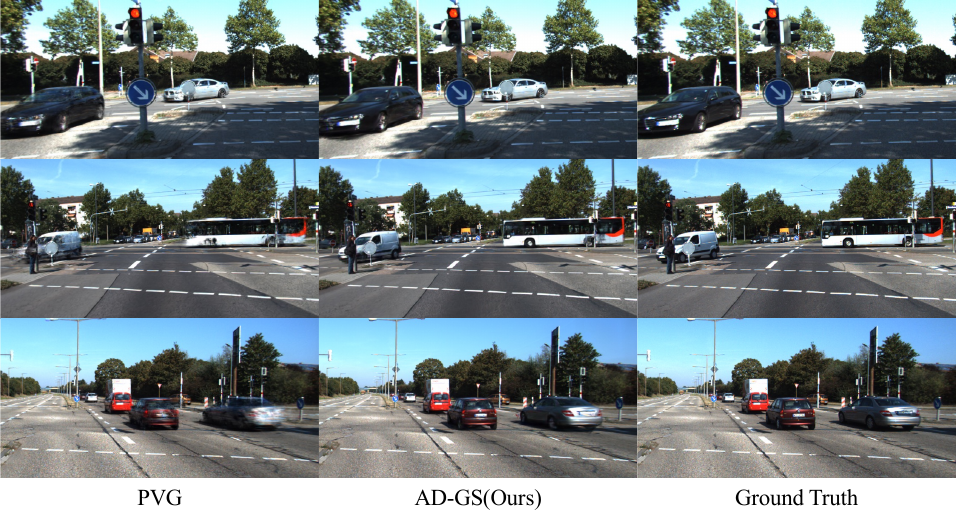}
    \caption{Additional qualitative comparisons on the KITTI~\cite{kitti} dataset.}
    \label{fig:supp-comp-kitti}
\end{figure*}

\begin{figure*}[t]
    \centering
    \includegraphics[width=\linewidth]{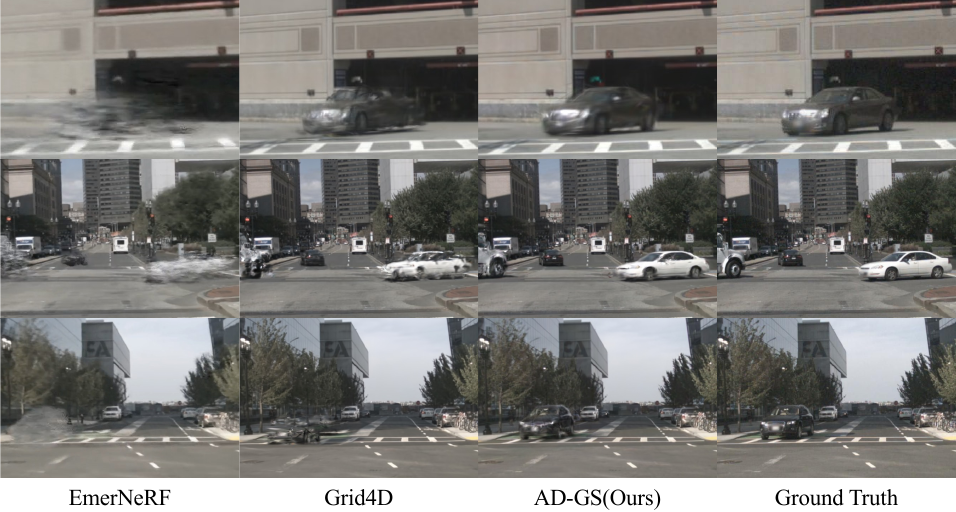}
    \caption{Additional qualitative comparisons on the nuScenes~\cite{nuscenes} dataset.}
    \label{fig:supp-comp-nuscenes}
\end{figure*}

\noindent\textbf{Additional Visualization.}
Figure~\ref{fig:supp-comp-kitti} shows the additional rendering results on the KITTI~\cite{kitti} dataset.
Figure~\ref{fig:supp-comp-waymo} is the additional rendering results on the Waymo~\cite{waymo} dataset.
Figure~\ref{fig:supp-comp-nuscenes} displays the additional rendering results in the nuScenes~\cite{nuscenes} dataset.

\section{Limitations}
Although \modelname~achieves state-of-the-art performances in self-supervised auto-driving scene rendering, it still has several limitations.
In some cases, \modelname~fails to outperform certain state-of-the-art rendering models that leverage manual 3D annotations to avoid the challenges of motion and object reconstruction.
Additionally, our model may produce artifacts if the quality of the pseudo-labels is quite low.
As illustrated in Figure~\ref{fig:supp-limitations}~(a), our model probably fails to reconstruct objects with highly complex motions and structures.
Moreover, when an object is only visible for an extremely brief period, such as about 10 frames, \modelname~might obtain suboptimal rendering results, as shown in Figure~\ref{fig:supp-limitations}~(b).

\end{document}